\documentclass[letterpaper]{article} 
\pdfoutput=1
\usepackage{aaai24}  
\usepackage{times}  
\usepackage{helvet}  
\usepackage{courier}  
\usepackage[hyphens]{url}  
\usepackage{graphicx} 
\usepackage{natbib}  
\usepackage{caption} 
\usepackage{algorithm}
\usepackage{algorithmic}

\usepackage{newfloat}
\usepackage{listings}
\DeclareCaptionStyle{ruled}{labelfont=normalfont,labelsep=colon,strut=off} 
\floatstyle{ruled}
\newfloat{listing}{tb}{lst}{}
\floatname{listing}{Listing}
\title{Exposing the Deception: Uncovering More Forgery Clues for Deepfake Detection}
\author{
    Zhongjie Ba\textsuperscript{\rm 1,\rm 2},
    Qingyu Liu\textsuperscript{\rm 1,\rm 2},
    Zhenguang Liu\textsuperscript{\rm 1,\rm 2} \footnote{The corresponding author.},
    Shuang Wu\textsuperscript{\rm 3},
    Feng Lin\textsuperscript{\rm 1,\rm 2},
    Li Lu\textsuperscript{\rm 1,\rm 2},
    Kui Ren\textsuperscript{\rm 1,\rm 2}
}
\affiliations{
    \textsuperscript{\rm 1}State Key Lab. of Blockchain and Data Security, Zhejiang University, Hangzhou, China\\
    \textsuperscript{\rm 2}ZJU-Hangzhou Global Scientific and Technological Innovation Center, Hangzhou, China\\
    \textsuperscript{\rm 3}Black Sesame Technologies, Singapore\\

    \{zhongjieba,qingyuliu\}@zju.edu.cn,
    liuzhenguang2008@gmail.com,
    wushuang@outlook.sg,
    \{flin,li.lu,kuiren\}@zju.edu.cn,
}

\usepackage{bibentry}

\usepackage{booktabs}
\usepackage{bbding}
\usepackage{multirow}
\usepackage{subfig}
\usepackage{amsmath}
\usepackage{amssymb}
\usepackage{soul} 
\usepackage{color, xcolor} 

\usepackage{threeparttable} 

\begin{document}

\maketitle

\begin{abstract}

Deepfake technology has given rise to a spectrum of novel and compelling applications. Unfortunately, the widespread proliferation of high-fidelity fake videos has led to pervasive confusion and deception, shattering our faith that seeing is believing. One aspect that has been overlooked so far is that current deepfake detection approaches may easily fall into the trap of overfitting, focusing only on forgery clues within one or a few local regions. Moreover, existing works heavily rely on neural networks to extract forgery features, lacking theoretical constraints guaranteeing that sufficient forgery clues are extracted and superfluous features are eliminated. These deficiencies culminate in unsatisfactory accuracy and limited generalizability in real-life scenarios.

\quad In this paper, we try to tackle these challenges through three designs: (1) We present a novel framework to capture broader forgery clues by extracting multiple non-overlapping local representations and fusing them into a global semantic-rich feature. (2) Based on the information bottleneck theory, we derive Local Information Loss to guarantee the orthogonality of local representations while preserving comprehensive task-relevant information. (3) Further, to fuse the local representations and remove task-irrelevant information, we arrive at a Global Information Loss through the theoretical analysis of mutual information. Empirically, our method achieves state-of-the-art performance on five benchmark datasets.
Our code is available at \url{https://github.com/QingyuLiu/Exposing-the-Deception}, hoping to inspire researchers.  

\end{abstract}

\section{1\quad Introduction}
Fueled by the accessibility of large-scale video datasets and the maturity of deepfake technologies~\cite{nirkin2019fsgan,li2019faceshifter}, one may effortlessly create massive forgery videos beyond human discernibility. However, malicious usage of deepfake can have serious influences, ranging from identity theft and privacy violations to large-scale financial frauds and dissemination of misinformation. For instance, in March 2022, hackers created a fake video of the Ukrainian president Zelenskyy in which he stands at a podium and addresses Ukrainian soldiers to lay down their arms. Such events are far from isolated, and they highlight the risk of deepfake technology in misleading the public and undermining trust. Consequently, accurate and effective deepfake detection is essential for mitigating these risks. 


\begin{figure}
	\centering
 \includegraphics[width=1\linewidth]{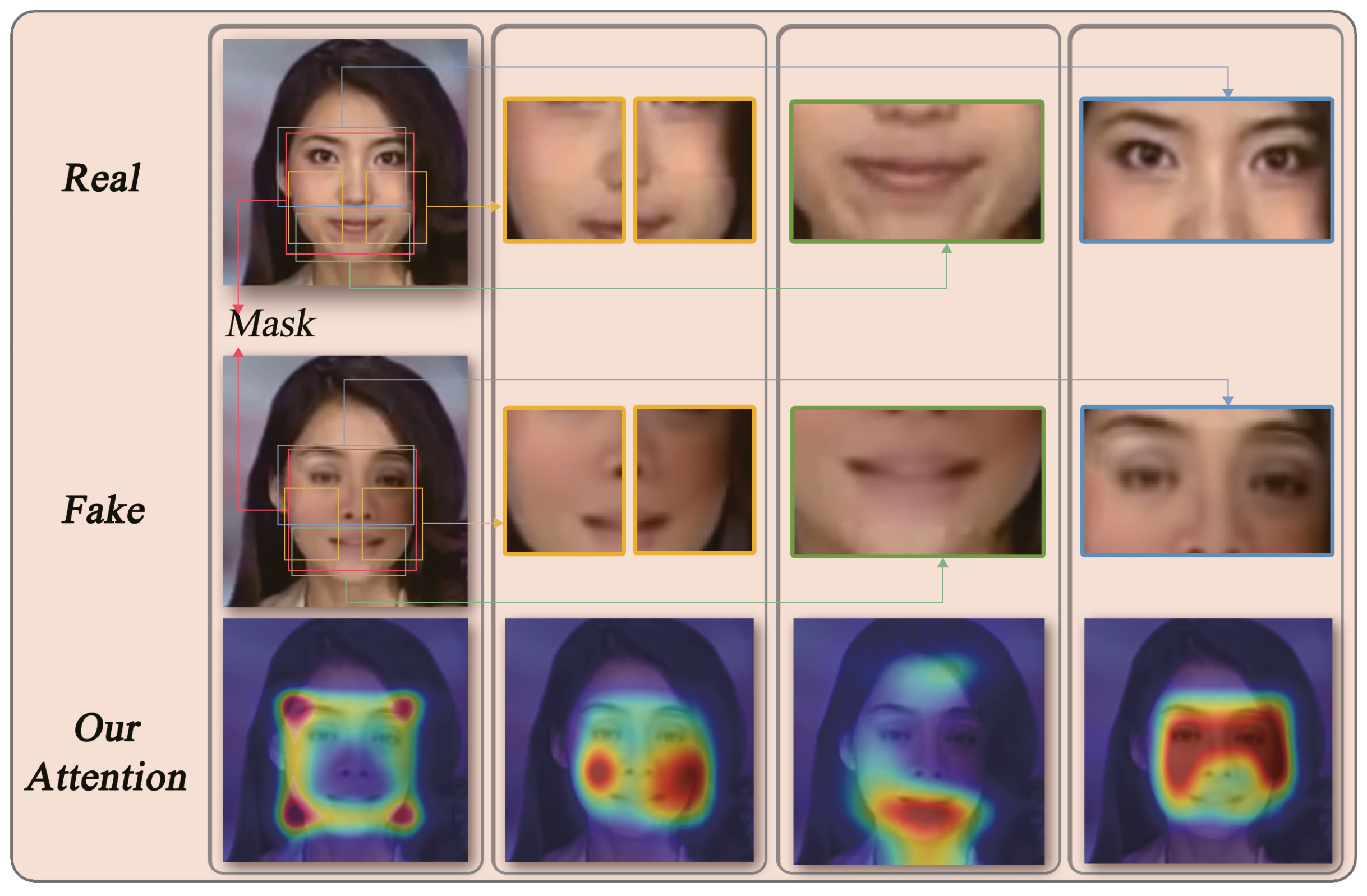}
	\caption{Example visualization of four local salient features obtained by our method. Each feature focuses on distinct forgery regions with little overlap. We zoom in to show the detailed differences for these regions between a real sample and a fake sample. Our method can grasp broader forgery clues including \emph{blending ghosts, consistent and symmetrical skin tones, tooth details}, and \emph{stitching seams}.}
	\label{fig:intro} 
\end{figure}
Fundamentally, deepfake detection amounts to recognizing forgery clues that distinguish real and synthetic images. 
There emerge many studies for deepfake detection, which can be roughly categorized into two main branches. One line of work~\cite{afchar2018mesonet,dolhansky2020deepfake} employs CNN networks to \textit{automatically} learn the clues within manipulated images. Another line of work is dedicated to pondering and exploring the differences between fake and real images by incorporating human observation and understanding. These approaches hone in on high-level semantic imperfections of counterfeits~\cite{haliassos2021lips}, as well as underlying imperceptible patterns in artifacts (such as \textit{blending ghost}~\cite{shiohara2022detecting} and \textit{frequency domain anomalies}~\cite{liu2021spatial}). 

After scrutinizing and experimenting with the implementations of state-of-the-art  approaches, we obtain two empirical insights: (1) 
Despite achieving a high AUC on the training dataset, 
current methods usually experience a substantial decrease in AUC on unseen datasets. This may stem from the fact that current methods tend to unintentionally learn shortcuts for the training dataset, focusing only on one or a few forgery clues.
(2) Current works heavily rely on neural networks to automatically extract forgery features, lacking rigorous theoretical guarantees to capture sufficient label-relevant information and to eliminate superfluous information. 
Consequently, the extracted features may converge to insufficient representations or trivial features, compromising the accuracy of such methods.
Motivated by these, we advocate extracting broader forgery clues for deepfake detection and seek to lay the mathematical foundation for sufficient forgery feature extraction. Specifically, we first adaptively extract multiple disentangled local features focused on non-overlapping aspects of the suspect image (as in Fig.~\ref{fig:intro}).  To ensure the orthogonality of these local features while preserving comprehensive task-relevant information, we utilize mutual information to derive an information bottleneck objective, \emph{i.e.,} Local Information Loss. 
Secondly, we fuse local features into a global representation guided by Global Information Loss that serves to eliminate task-irrelevant information. 

To evaluate the effectiveness of our method, we conduct extensive experiments on five widely used benchmark datasets, \emph{i.e.}, FaceForensics++~\cite{rossler2019faceforensics++}, two versions of Celeb-DF~\cite{li2020celeb}, and two versions of DFDC~\cite{dolhansky2020deepfake}. We also conduct ablation studies to assess the efficacy of each key component in our method. Our method achieves state-of-the-art performance for both in-dataset (the training and test datasets are sampled from the same domain) and cross-dataset (the training and test datasets are two different datasets) settings. In summary, our contributions are as follows:
\begin{itemize}
    \item We propose a novel framework for deepfake detection that aims to obtain broader forgery clues.
    \item We mathematically formulate a mutual information objective to effectively extract disentangled task-relevant local features. Additionally, we introduce another objective for aggregating the local features and eliminating superfluous information. We provide a rigorous theoretical analysis to show how these mutual information objectives can be optimized.
    \item Empirically, our method achieves state-of-the-art performance on five benchmark datasets. Interesting and new insights are also presented (\emph{e.g.,} most deepfake detection approaches tend to focus on only a few specific regions around the face swap boundaries).  

\end{itemize}

\section{2\quad Related Work}
Fueled by the maturity of deep learning models and large-scale labeled datasets, deep learning has found its applications in various fields \cite{zhang2023text,wei2023imitation,liu2022copy,chiou2020zero,liu2023rethinking,song2023relation}, especially for deepfake. The ease of access and misuse of deepfake technology has led to the materialization of severe risks, and developing deepfake detection to counteract such threats is all the more pertinent and urgent. Deepfake detection~\cite{ying2023bootstrapping,ba2023transferring,hua2023tip,wu2023generalizing,pan2023dfil,shuai2023locate} faces a significant challenge posed by the sophistication of deepfake technology that can create highly realistic content that is barely distinguishable from real ones.

A large body of literature~\cite{dong2022protecting,li2020face,shiohara2022detecting,chen2022self,haliassos2021lips,zhao2021multi} focuses on semantic facial feature clues of forgeries. ICT~\cite{dong2022protecting} models identity differences in the inner and outer facial regions. Face X-ray~\cite{li2020face} and SBIs~\cite{shiohara2022detecting} find the blending boundaries of face swap as evidence for forged images and build private augmented datasets. Chen.\textit{et al.}~\cite{chen2022self} further expands upon blending-based forgeries, considering the eyes, nose, mouth, and blending ratios. LipForensics~\cite{haliassos2021lips} observes the irregularities of mouth movements in forgery videos. However, such methods only apply to the detection of face swaps and semantic-guided forgery detection cannot be exhaustive vis-à-vis the rapid development of deepfake techniques. 

Another class of works~\cite{frank2020leveraging,liu2021spatial,luo2021generalizing,qian2020thinking} proposes to take into further consideration human understanding differences in the frequency domain. Qian.\textit{et al.}~\cite{qian2020thinking} employ frequency as complementary evidence for detecting forgeries, which can reveal either subtle forgery clues or compression errors. Frank.\textit{et al.}~\cite{frank2020leveraging} and SPSL~\cite{liu2021spatial} search for ghost artifacts resulting from up-sampling operations in generative networks. While such works include more features than pure semantically visual clues, the additional features modelled tend to be domain-specific, thereby failing to generalize well to cross-dataset scenarios.

Researchers have also engaged in multi-headed attention modules to correlate the low-level textural features and high-level semantics at different regions for deepfake detection~\cite{zhao2021multi}. Nevertheless, a challenge persists, as there exists no concrete theoretical assurance that these attention regions segmented based on the paradigm of human vision remain entirely task-relevant and independent. Furthermore, the performance of such attention-based models is greatly affected by data scarcity.
\section{3\quad Methodology}
\subsection{3.1\quad Overview}
\begin{figure*}[!ht]
  \centering
  \includegraphics[width=0.85\linewidth]{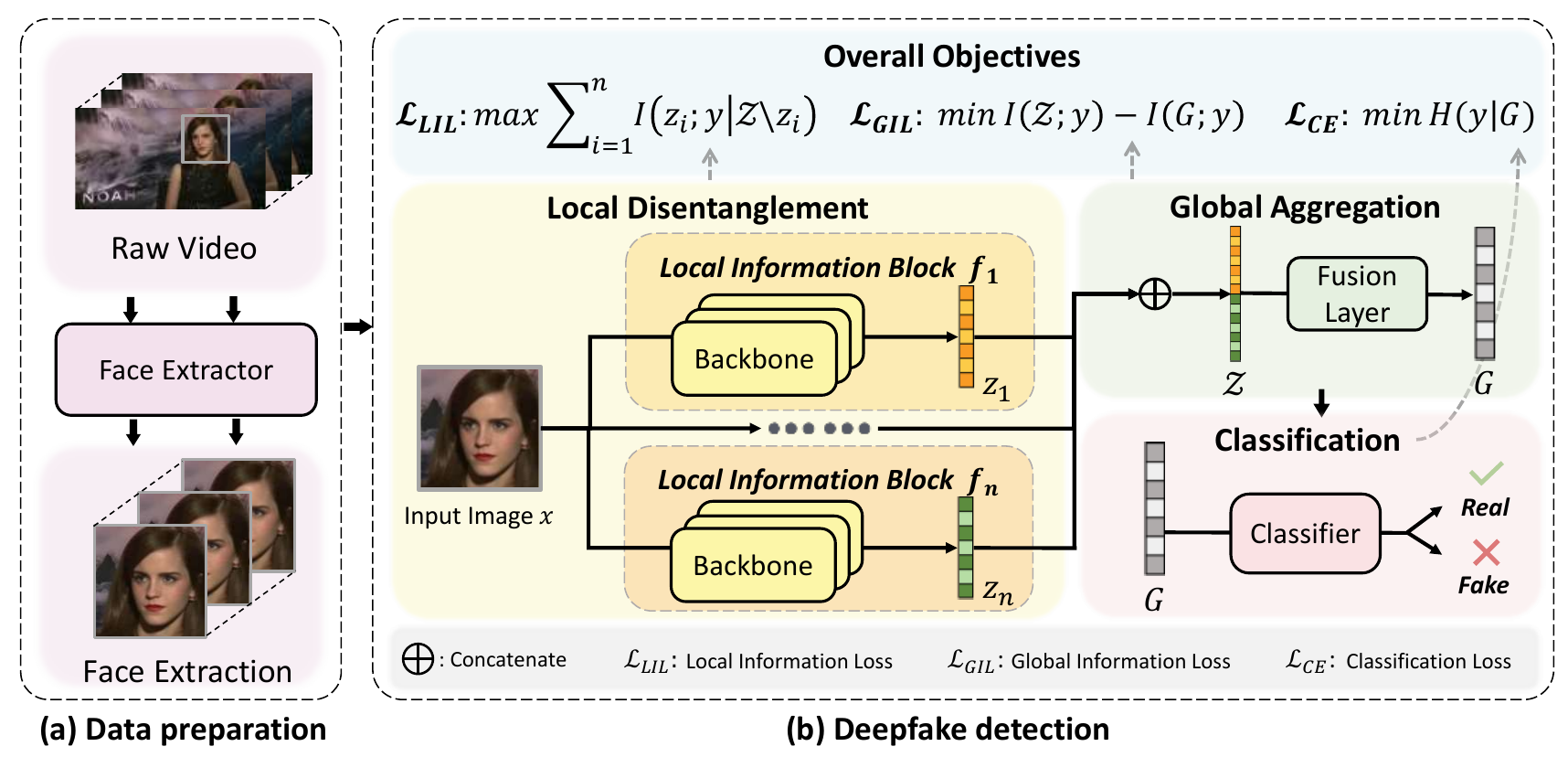}
  \caption{Method overview. In the data preparation phase, we first extract frame-level facial bounding boxes from raw videos. For deepfake detection, our method consists of three modules. i) We first employ local information blocks $f_i$ to extract multiple disentangled local features $z_i$ corresponding to different forgery regions. We introduce local information loss to ensure that $z_i$ has comprehensive forgery-related information and is orthogonal to $z_j$. ii) We fuse all $z_i$  into a global feature $G$ under the guidance of a Global Information Loss. iii) Finally, $G$ is passed to the classification module to output the prediction result. We design our Local and Global Information Loss based on information bottleneck theory.}
  \label{fig:framework}
\end{figure*}

Presented with a suspect image, we aim to judge its authenticity by extracting forgery clues that could distinguish between genuine and synthetic images. Technically, deepfake detection can be viewed as a binary classification problem.

Early methods~\cite{afchar2018mesonet,dolhansky2020deepfake} directly utilize deep neural networks to automatically learn differences between genuine and synthetic images. Recent works~\cite{shiohara2022detecting,haliassos2021lips,liu2021spatial} try to draw inspiration from human understanding and explore human-perceivable forgery clues. 
Unfortunately, current approaches tend to unintentionally learn shortcuts\footnote{Shortcuts are decision rules optimized for benchmark performance but incapable of transferring to more challenging testing conditions due to a domain gap.}, which actually makes approaches focus only on one or a few forgery regions. This overfitting issue manifests as the limited generalizability of state-of-the-art methods, namely significant accuracy decreases when applied to unseen datasets. Furthermore, due to the lack of rigorous theoretical constraints, neural networks of current methods may converge to trivial features or insufficient representations.

Motivated by these, we propose to broaden our extraction of forgery clues by adaptively extracting multiple non-overlapping local features. We also establish a theoretical foundation to ensure the orthogonality and sufficiency of extracted features. 
Overall, our approach consists of two key components: the \emph{local disentanglement module} and the \emph{global aggregation module}. The local disentanglement module serves to extract the non-overlapping local features while the global aggregation module is designed to aggregate them into a global representation.



The pipeline of our proposed approach is shown  in Fig.~\ref{fig:framework}, which can be summarized as follows. (1) For image preprocessing, we extract face regions using a popular pre-trained backbone network. (2) Given the preprocessed image as an input, we design the local disentanglement module to extract multiple local features. The local disentanglement module comprises $n$ local information blocks $\{f_i\}_{i=1}^n$, each extracting a local feature $z_i$. $f_i$ consists of feature extraction backbone networks such as ResNet~\cite{he2016deep}. To ensure that local features contain comprehensive information related to the task and $z_i$ is orthogonal to $z_j (i\neq j)$, we derive the Local Information Loss $\mathcal{L}_{LIL}$. (3) Thereafter, we design the global aggregation module to fuse local features. Specifically, we first concatenate all local features to a joint local representation $\mathcal{Z}=\bigoplus_i^n z_i$. Then, a fusion layer $f_g$ serves to fuse and compress $\mathcal{Z}$ to obtain our final global representation $G$ for classification. To guide this global representation extraction, we design a Global Information Loss $\mathcal{L}_{GIL}$ that facilitates the retaining of sufficient task-related information and the elimination of superfluous information in $\mathcal{Z}$. 

In what follows, we elaborate on the details of the \textit{local disentanglement} and \textit{global aggregation} modules one by one.
\subsection{3.2\quad Local Disentanglement Module}
\label{sec:lil}
In this section, we provide the key derivation of Local Information Loss within the local disentanglement module.

\begin{figure}[!ht]
	\centering
	\subfloat[\label{fig:vene_local}Optimizing local features]{
		\includegraphics[width=\linewidth]{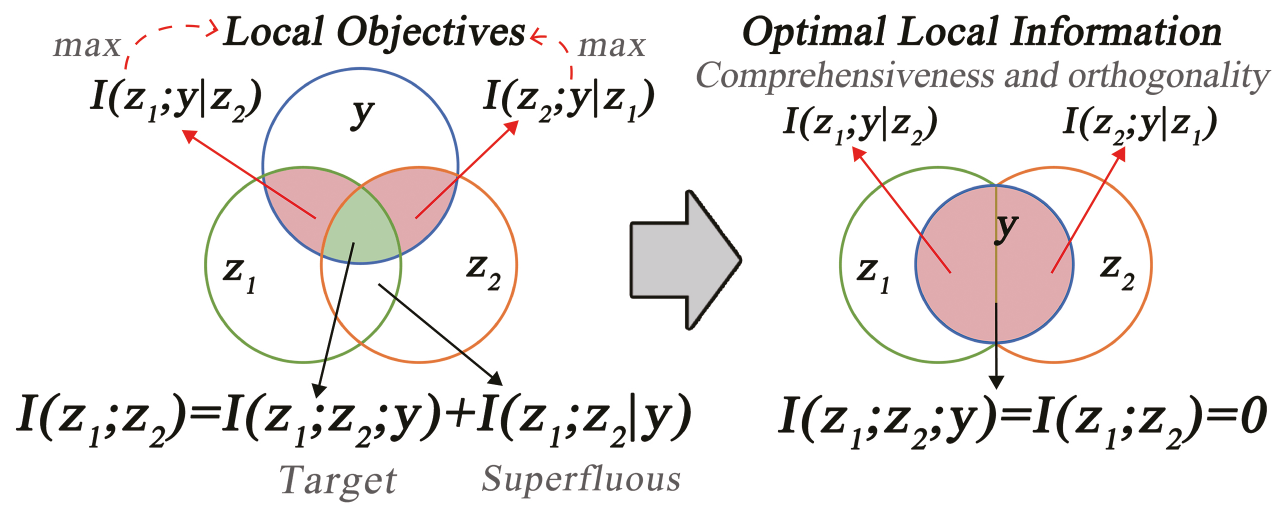}}
	\\
	\subfloat[\label{fig:vene_global}Optimizing the global feature]{
		\includegraphics[width=0.9\linewidth]{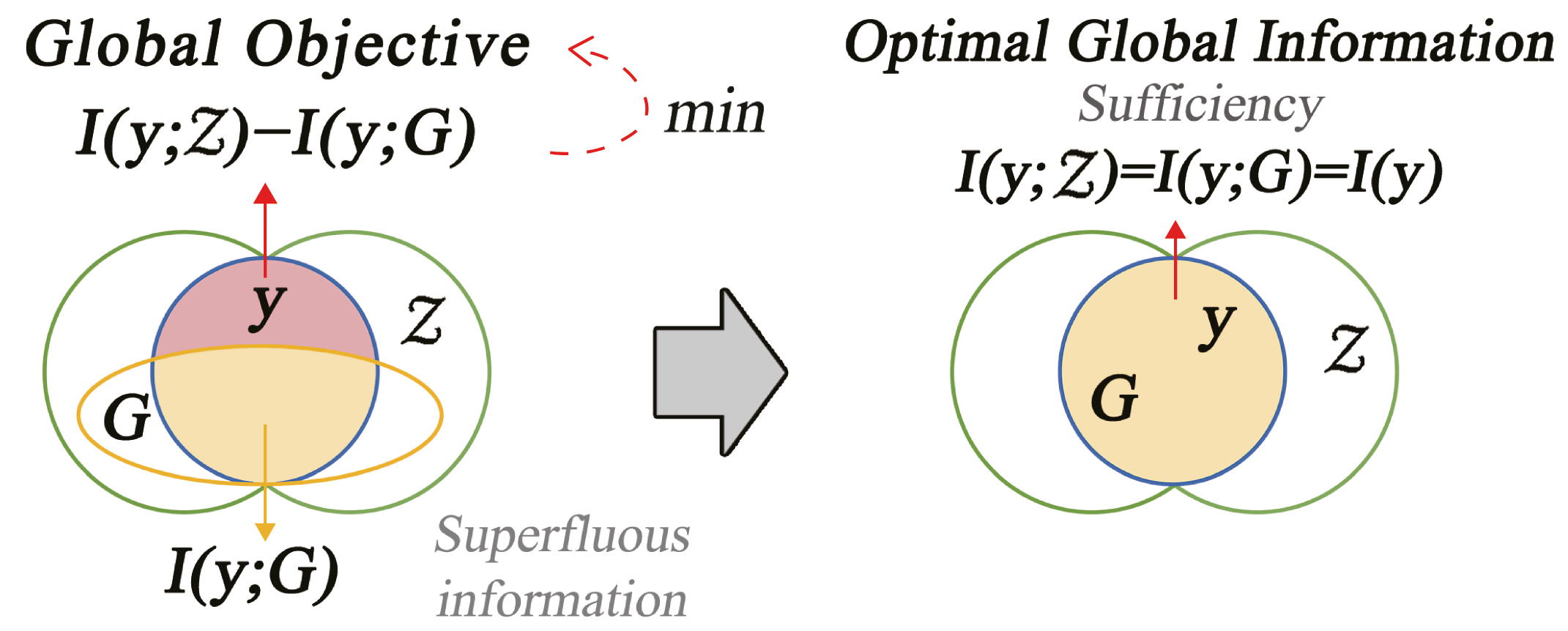}}
	\caption{Information content of feature representations}
	\label{fig:vene}
\end{figure}

Given an input image $x$ with $n$ ($n\geq 2$) associated local feature representations $z_i$, our Local Information Loss objective seeks to ensure two fundamental properties within the joint local representation $\mathcal{Z}=\bigoplus_i^n z_i$, \emph{i.e.,} \textbf{comprehensiveness and orthogonality}. Comprehensiveness mandates the inclusion of maximal task-relevant information within $\mathcal{Z}$, while orthogonality necessitates that the individual local features $z_i$ remain non-overlapping. To facilitate understanding, Fig.~\ref{fig:vene} shows the information relationship when $n=2$. 

In the terminology of mutual information theory, the relationship between labels $y$ and $\mathcal{Z}$ is expressed as:
\begin{equation}
\small
\begin{aligned}
  I ( y;\mathcal{Z} )=I( y ;z_1, \cdots, z_n)=H(y)-H ( y\mid \mathcal{Z}),
\end{aligned}
\end{equation}
where $I(*)$ is mutual information and $H(*)$ is entropy.  $I ( y;\mathcal{Z} )$ expresses the amount of predictive information (\emph{i.e.}, current task-related information) contained in $\mathcal{Z}$. $H ( y\mid \mathcal{Z} )$ and $H(y)$ represent the required and whole information related to the task, respectively. The comprehensiveness objective of information in $\mathcal{Z}$  is given by:
\begin{equation}
\small
\begin{aligned}
\label{eq:y_Z}  \max I( y ;\mathcal{Z} ).
\end{aligned}
\end{equation}

The orthogonality condition between two probability distributions is equivalent to them having zero mutual information. As such, we can disentangle local feature representations by minimizing the mutual information between them, \emph{i.e.,} $\min \sum_{i \neq j}^n I(z_i; z_j).$
According to the definition of interaction information~\cite{mcgill1954multivariate}, $I(z_i;z_j)$ can be further decomposed into:
\begin{equation}
\small
\begin{aligned}
\label{eq:zi_zj_2}
I(z_i; z_j)=\underbrace{I ( z_i ; z_j;y )}_\text{target}+\underbrace{I ( z_i ;z_j \mid y )}_\text{superfluous},
\end{aligned}
\end{equation}
where $I( z_i;z_j ; y)$ represents the amount of label information retained within both $z_i$ and $z_j$, while $I ( z_i;z_j \mid y )$ is extraneous (superfluous) information encoded within both $z_i$ and $z_j$, which is irrelevant to the task. 
For the orthogonality of local features, we are primarily concerned with label-related (target) information. As for the elimination of superfluous information, we formulate an objective inspired by the information bottleneck, namely Global Information Loss (which will be discussed later in the following section).
We first focus on the target term in Eq.~\ref{eq:zi_zj_2}, \textit{i.e.,}: 
\begin{equation}
\small
\begin{aligned}
\label{eq:zi_zj_y}  \min \sum_{i \neq j}^{n} I(z_i; z_j ; y ).
\end{aligned}
\end{equation}

By applying the chain rule for mutual information, $I(y;\mathcal{Z})=\sum_{i=1}^{n} I(z_i;y\mid z_1,\cdots,z_{i-1})$, we can rewrite Eq.~\ref{eq:y_Z} as:
\begin{equation}
\small
\begin{aligned}
\label{eq:y_Zexpanded}
\max I ( y; \mathcal{Z}) \leq \max \sum_{i \neq j}^{n} {I( z_i;y\mid \mathcal{Z} \setminus z_i ) +I(z_i; z_j ; y )},
\end{aligned}
\end{equation}
where $\mathcal{Z} \setminus z_i \equiv z_1 \oplus \cdots \oplus z_{i-1} \oplus z_{i+1} \oplus \cdots \oplus z_n$. Overall, the comprehensiveness and orthogonality constraints for local feature extraction can be achieved by simultaneously optimizing Eq.~\ref{eq:y_Zexpanded} and Eq.~\ref{eq:zi_zj_y}. It is worth noting that these optimization objectives are in conflict with $I(z_i; z_j ; y )$. After resolving these conflicting constraints,  the local objective is eventually:
\begin{equation}
\small
\begin{aligned}
\label{eq:LIL} \max \sum_{i=1}^{n} I( z_i;y\mid \mathcal{Z} \setminus z_i ).
\end{aligned}
\end{equation}
Intuitively, the local objective corresponds to the red regions illustrated in Fig.~\ref{fig:vene_local}. By optimizing Eq.~\ref{eq:LIL}, our goal is to ideally cover all task-relevant information with disentangled local features.

{However, directly estimating Eq.~\ref{eq:LIL} is intractable in general. Earlier works~\cite{poole2019variational} have pointed out major difficulties in mutual information estimation, primarily due to the curse of dimensionality (the amount of samples for accurately estimating mutual information scales exponentially with the embedding dimension). In light of this, we optimize Eq.~\ref{eq:LIL} via a variational approach instead of explicitly estimating the mutual information. 
We have the following theorem (detailed proof is in supplementary files):

\textbf{Theorem:} \quad\emph{Eq.~\ref{eq:LIL} has a lower bound due to:}
\begin{equation}
\small
\begin{aligned}
 \sum_{i=1}^{n} I( z_i;y\mid \mathcal{Z} \setminus z_i ) \geq \sum_{i=1}^{n} D_{KL}\left[\mathbb{P}_z \| \mathbb{P}_{z \setminus z_i}\right],
\end{aligned}
\end{equation}
where $\mathbb{P}_{\mathcal{Z} \setminus z_i} = p \left( y\mid \mathcal{Z} \setminus z_i \right), \mathbb{P}_\mathcal{Z} = p \left(y\mid \mathcal{Z} \right)$ represent the predicted
distributions. $D_{K L}$ denotes the Kullback-Leibler (KL) divergence.  

Given the above analytical derivations, we can thus formulate the Local Information Loss as:
\begin{equation}
\small
\begin{aligned}
\label{eq.lossLIL}
\mathcal{L}_{L I L}=\min _\theta \exp \left( -\sum_{i=1}^{n} D_{KL}\left[\mathbb{P}_z \| \mathbb{P}_{z \setminus z_i}\right]\right),
\end{aligned}
\end{equation}
where $\theta$ denotes the model parameters in our local disentanglement module. Here, since the KL-divergence is not bounded above, \emph{i.e.} $D_{KL}\in [0,\infty)$, we take the exponential of its negative value to transform the objective from maximization to minimization. The transformed objective is bounded within $(0,1]$ which is numerically advantageous. Upon optimizing for this objective, local features are constrained to be mutually orthogonal while simultaneously approaching the maximal covering of all task-related information. In this way, our method uncovers more forgery clues and disentangles forgery regions adaptively, thus obtaining feature representations with richer task-related information.
\subsection{3.3\quad Global Aggregation Module} 
Following the local disentanglement module, the concatenated local features $\mathcal{Z}$ encompass comprehensive but not purified information related to the task. Thus, we pass  $\mathcal{Z}$ through our global aggregation module, which plays the role of an information bottleneck\footnote{The concept of information bottlenecks is proposed in \cite{tishby2000information} which attributes the robustness of a machine learning model to its ability to distill superfluous noises while retaining only useful information.} to eliminate the superfluous information and obtain a global representation $G$. The information bottleneck objective can be formulated as:
\begin{equation}
\small
\begin{aligned}
\label{eq:IB}
\mathcal{L}_{IB} = H(G) - I (y;G),
\end{aligned}
\end{equation}
where $H(G)$ denotes the total information content in $G$. Once again, estimating $\mathcal{L}_{IB}$ is an intractable problem in practice due to the curse of dimensionality. Minimizing superfluous information will therefore be delegated to the network operations and is not explicitly supervised.

Therefore, to ensure that $G$ has sufficient label information, we employ a variational approach once again. Since $G$ is a representation learnt from $\mathcal{Z}$, the task-relevant information in $G$ is upper-bounded by that in $\mathcal{Z}$, denoted as $I(y;G) \leq I(y; \mathcal{Z})$. By minimizing the label information difference between local features and the global feature, we optimize $G$ for the \textbf{sufficiency} of label information:
\begin{equation}
\small
\begin{aligned}
\label{eq:g_z} \min I ( y;\mathcal{Z})-I ( y;G).
\end{aligned}
\end{equation}

\noindent We make use of the following theorem from \cite{tian2021farewell}:
\begin{equation}
\small
\begin{aligned}
\min I(y;\mathcal{Z}) - I(y; G) \iff \min [D_{K L}[\mathbb{P}_\mathcal{Z} \| \mathbb{P}_G]]].
\end{aligned}
\end{equation}
Finally, we arrive at the Global Information Loss:
\begin{equation}
\small
\begin{aligned}
\label{eq.lossGIL}
\mathcal{L}_{G I L}=\min _{\phi} \mathbb{E}_{G \sim E_\phi(G \mid \mathcal{Z})}\left[D_{K L}[\mathbb{P}_\mathcal{Z} \| \mathbb{P}_G]\right],
\end{aligned}
\end{equation}
where $\phi$ denotes the model parameters of the global aggregation module.}

\textbf{Overall Objective} \quad The overall objective for our framework consists of a cross-entropy classification loss, Local Information Loss, and Global Information Loss:
\begin{equation}
\small
\begin{aligned}
\label{eq:total_loss}
\mathcal{L}=\mathcal{L}_{CE}+\alpha \mathcal{L}_{LIL}+\beta \mathcal{L}_{G I L},
\end{aligned}
\end{equation}
where $\alpha$ and $\beta$ are hyperparameters for the model.
\section{4\quad Evaluation}
In this section, we conduct extensive experiments on five large-scale deepfake datasets to evaluate the proposed method, including setup, comparison with state-of-the-art methods, ablation study, and visualization results. See the supplementary file for more experimental results.

\subsection{4.1\quad Experimental Setup}

\textbf{Datasets.} Following existing deepfake detection approaches~\cite{chen2022self,bai2023aunet}, we evaluate our model on five public datasets, namely FaceForensics++ (FF++)~\cite{rossler2019faceforensics++}, two versions of Celeb-DF~\cite{li2020celeb} and two versions of DeepFake Detection Challenge (DFDC)~\cite{dolhansky2020deepfake} datasets. \textbf{FF++} dataset, which is the most widely used dataset, utilizes four forgery-generation methods for producing 4,000 forgery videos, \emph{i.e.,} DeepFakes (DF), Face2Face (FF), FaceSwap (FS), and  NeuralTextures (NT). FF++ has three compression versions and we use the high-quality level (C23) one for training. Celeb-DF dataset contains two versions, termed \textbf{Celeb-DF-V1} (CD1) and \textbf{Celeb-DF-V2} (CD2). CD1 consists of 408 pristine videos and 795 manipulated videos, while CD2 contains 590 real videos and 5,639 DeepFake videos. DFDC dataset includes \textbf{DFDC-Preview} (DFDC-P) and \textbf{DFDC}. DFDC-P as the preview of DFDC consists of 5,214 videos. DFDC, as one of the most large-scale face swap datasets, contains more than 110,000 videos sourced from 3,426 actors. 

\textbf{Implementation details.} In data pre-processing, we use the state-of-the-art face extractor RetinaFace~\cite{deng2020retinaface} and oversample pristine videos to balance training datasets. For the model architecture, we employ four local information blocks (LIBs) using pre-trained ResNet-34~\cite{he2016deep} as the backbone. For training, we use the method in ~\cite{liebel2018auxiliary} to determine $\alpha$ and $\beta$ in Eq.~\ref{eq:total_loss}, to automatically balance the weights for these loss terms. 

\textbf{Evaluation metrics.} We utilize the Accuracy (ACC), Area Under Receiver Operating Characteristic Curve (AUC), and log-loss score for empirical evaluation. (1) \textbf{ACC}. We employ the accuracy rate as one of the metrics in our evaluations, which is commonly used in deepfake detection tasks. (2) \textbf{AUC}. Considering the imbalance of pristine and manipulated videos in the datasets, we use AUC as the predominant evaluation metric. (3) \textbf{Logloss}. This is the evaluation metric designated for the Deepfake Detection Challenge. We evaluate the log-loss score to benchmark our method against winning teams. 
By default, we use frame-level metrics. Since our method uses a single frame as input, we also compute video-level AUC (as in ~\cite{haliassos2021lips}) for a more comprehensive comparison with video-level detection methods. 

\subsection{4.2\quad Comparison with Existing Methods}

\begin{table*}[h!]
\centering
\small
\begin{tabular}{lc|lc|lcc}
\toprule
\multicolumn{2}{c|}{FF++(C23)}                              & \multicolumn{2}{c|}{Celeb-DF-V2}                           & \multicolumn{3}{c}{DFDC}             \\ 
[0.7ex]\hline
Method                   & AUC$\uparrow$                    & Method                         & AUC$\uparrow$   & Method             & AUC$\uparrow$   & LogLoss$\downarrow$ \rule[-1.2ex]{0ex}{1ex}\\[1.ex]\hline
Xception~\cite{rossler2019faceforensics++}                   & 0.963                  & DeepfakeUCL~\cite{fung2021deepfakeucl}            & 0.905 & Selim Seferbekov$^*$   & 0.882 & 0.4279  \\
Xception-ELA
&
  0.948 &
  SBIs~\cite{shiohara2022detecting} &
  0.937 & NTechLab$^*$           & 0.880 & 0.4345
   \\
SPSL~\cite{masi2020two} & 0.943                  & Agarwal et al. 2020
& 0.990 &  Eighteen Years Old$^*$ & 0.886 & 0.4347 \\
Face X-ray~\cite{li2020face} & 0.874                  &                         Wu et al. 2023  
& \underline{0.998} &  WM$^*$                 & 0.883 & 0.4284  \\
TD-3DCNN~\cite{zhang2021detecting}                 & 0.722 & TD-3DCNN                & 0.888 & TD-3DCNN            & 0.790 & \underline{0.3670}  \\
  F$^3$-Net~\cite{qian2020thinking}             & \underline{0.981} & Xception                              & 0.985  &  Chugh et al. 2020
  &
  \underline{0.907} &
  - \\
FInfer~\cite{hu2022finfer}                       & 0.957 & FInfer                   & 0.933  & FInfer              & 0.829 & -       \\
 [0.7ex]\hline
\textbf{Ours (ResNet34)} &
\textbf{0.983} &
 {\textbf{Ours (ResNet34)}}  &
 
  \textbf{0.999} &
  {\textbf{Ours (ResNet34)}} &
  {\textbf{0.939}} &
  \textbf{0.3379} \\ \bottomrule
\end{tabular}
\caption{In-dataset comparison results on FF++, Celeb-DF-V2, and DFDC. We train and test models on the same dataset, reporting the frame-level AUC and LogLoss. $^*$ is the method of winning the top four teams in DFDC. The bold and underline mark the best and second performances, respectively.}
\label{tab:in-dataset}
\end{table*}

\begin{table*}[h!]

\centering

\small
\begin{tabular*}{0.8\textwidth}{@{\extracolsep{\fill}} lccccc}
\toprule

\
Method &
\begin{tabular}[c]{@{}c@{}}Training\\ dataset\end{tabular} &
  {CD1} &
  {CD2} &
  {DFDC-P} &
  DFDC \\  \midrule
Xception~\cite{rossler2019faceforensics++}        & FF++ & 0.750$^*$      & 0.778$^*$      & 0.698$^*$      & 0.636$^*$      \\
DSP-FWA~\cite{li2018exposing}         & FF++ & 0.785$^*$      & 0.814$^*$      & 0.595$^*$      &-          \\
Meso4~\cite{afchar2018mesonet}           & FF++ & 0.422$^*$      & 0.536$^*$      & 0.594$^*$      &-         \\
F$^3$-Net~\cite{qian2020thinking}          & FF++ &-         & 0.712$^*$      & 0.729$^*$      & 0.646$^*$      \\
 Face X-ray~\cite{li2020face}      & PD   & 0.806       & 0.742$^*$      & 0.809       &-             \\
Multi-Attention~\cite{zhao2021multi} & FF++ &-         & 0.674       &-         & 0.680$^*$  \\
OST~\cite{chen2022ost}             & FF++ & 0.748       &-         & \underline{0.833} &-        \\
HCIL~\cite{gu2022hierarchical}            & FF++ &-         & 0.790       & 0.692       &-         \\
LiSiam~\cite{wang2022lisiam}          & FF++ & 0.811       & 0.782       &-         &-         \\
RECCE~\cite{cao2022end}           & FF++ &-         & 0.687       &-         & \underline{0.691} \\
ICT~\cite{dong2022protecting}             & PD   & \underline{0.814} & \underline{0.857} &-         &-         \\
DCL~\cite{sun2022dual}             & FF++ &-         & 0.823       & 0.767       &-         \\
IID~\cite{huang2023implicit}             & FF++ &-         & 0.838       & 0.812       &-         \\  \midrule
\textbf{Ours (ResNet-34)} &
  FF++ &
  \textbf{0.818} &
  \textbf{0.864} &
  \textbf{0.851} &
  \textbf{0.721} \\\bottomrule
\end{tabular*}

\caption{Cross-dataset comparison results (frame-level AUC) on Celeb-DF-V1 (CD1), Celeb-DF-V2 (CD2), DFDC-Preview (DFDC-P), and DFDC. We train our method on FF++ (C23) and test it on other benchmark datasets. The 'PD' means private data. * is collected from \cite{dong2022protecting,cao2022end,sun2022dual}, and other results are directly cited from the corresponding original paper. The bold and underline mark the best and second performances, respectively.}
\label{tab:cross-dataset}
\end{table*}

\begin{table*}[h!]

\centering

\small
\begin{tabular*}{0.8\textwidth}{@{\extracolsep{\fill}} lcccc}
\toprule
  Method &
  \begin{tabular}[c]{@{}c@{}}Training\\ dataset\end{tabular} &
  {CD2} &
  {DFDC-P} &
  {DFDC}  \\  \midrule
Xception~\cite{rossler2019faceforensics++}        & FF++ &  0.737$^*$      & 0.679$^*$      & 0.709$^*$      \\
F$^3$-Net~\cite{qian2020thinking}          & FF++      & 0.757$^*$      &-         & 0.709$^*$      \\
PCL+I2G~\cite{zhao2021learning}         & PD   & 0.900       & 0.744       & 0.675       \\
FST-Matching~\cite{dong2022explaining}    & FF++          & 0.894       &-         &-         \\
LipForensics~\cite{haliassos2021lips}    & FF++             & 0.824       &-         & 0.735       \\
FTCN~\cite{zheng2021exploring}            & FF++            & 0.869       & 0.740       & 0.710$^*$      \\
Luo.et al.~\cite{luo2021generalizing}      & FF++          &-         & 0.797       &-         \\
ResNet-34+ SBIs~\cite{shiohara2022detecting} & PD             & 0.870       & 0.822       & 0.664       \\
EFNB4+ SBIs~\cite{shiohara2022detecting}     & PD              & \underline{0.932}       & \underline{0.862} & 0.724       \\
RATF~\cite{gu2022region}             & FF++          & 0.765       & 0.691       &-         \\
Li.et al.~\cite{li2022wavelet}       & FF++           & 0.848       & 0.785       &-         \\
AltFreezing~\cite{wang2023altfreezing}     & FF++            & 0.895       &-         &-         \\
AUNet~\cite{bai2023aunet}           & PD            & 0.928 & \underline{0.862} & \underline{0.738} \\  \midrule
  {\textbf{Ours (ResNet-34)}} &
  {FF++} &
  \textbf{0.936} &
  \textbf{0.902} &
  \textbf{0.754} \\\bottomrule
\end{tabular*}

\caption{Cross-dataset comparison results (video-level AUC) on Celeb-DF-V2 (CD2), DFDC-Preview (DFDC-P), and DFDC. We train our method on FF++ (C23) and test it on other benchmark datasets. The 'PD' means private data. * is collected from \cite{shiohara2022detecting,bai2023aunet}, and other results are directly cited from the corresponding original paper. The bold and underline mark the best and second performances, respectively.}
\label{tab:cross-dataset-video}
\end{table*}

In this section, we benchmark our method against state-of-the-art deepfake detection methods for \emph{in-dataset} and \emph{cross-dataset} settings. 

\textbf{In-dataset performance}\quad In in-dataset evaluations, we train and test methods on FF++ (C23), CD2, and DFDC, respectively. Tab.~\ref{tab:in-dataset} presents in-dataset comparison results. Considering that most current deepfake detection methods have not yet released their codes, we directly cited their results in the corresponding original papers. 
From Tab.~\ref{tab:in-dataset}, we observe that our method is capable of consistently outperforming existing methods on all three benchmarks. For example, the AUC of our method is 0.939 on DFDC while the state-of-the-art detection method Chugh.\textit{et al.}~\cite{chugh2020not} is 0.907. The Logloss of our method is also outperformed by the champion team in DFDC.

\textbf{Cross-dataset Performance and Model Generalizability} \quad The cross-dataset setting is more challenging than the in-dataset setting for deepfake detection. To evaluate the generalization abilities of the methods on unseen datasets, we train the models on the FF++ (C23) dataset and test them on CD1, CD2, DFDC-P, and DFDC datasets. Since our method uses a single frame as input, in addition to frame-level comparisons, we also compute the average AUC over frames in a video for comparison with video-level methods. Tab.~\ref{tab:cross-dataset} and Tab.~\ref{tab:cross-dataset-video} demonstrate cross-dataset comparison results in terms of frame-level and video-level AUC, respectively. Our first insight is that state-of-the-art deepfake detection methods still suffer from relatively low AUC on unseen datasets, which reveals that such methods are prone to overfitting the training dataset. The second insight is that our method is more robust, with significant improvement when tested on unseen datasets. This reflects that our model has a better capability for uncovering forgery clues. The improvements in generalizability can be attributed to the information bottleneck in our framework design, where our model demonstrates a better capacity for identifying different forms of deepfake artifacts instead of merely the instances specific to the training dataset.
Overall, our method achieves state-of-the-art frame-level and video-level generalization performance. For frame-level comparisons, our method attains 0.818 and 0.857 AUCs on CD1 and CD2 respectively, outperforming the current state-of-the-art method ICT. Our method also improves the AUC on DFDC-P from 0.833 (OST) to 0.851, and on DFDC from 0.691 (RECCE) to 0.721. 
In contrast with video-level methods, our method surpasses the current state-of-the-art technique, AUNet, in terms of AUC with scores of 0.936, 0.902, and 0.754 on CD2, DFDC-P, and DFDC, respectively. Remarkably, despite employing solely traditional data augmentation techniques, our approach attains state-of-the-art performance across all four benchmarks, surpassing models (such as AUNet, SBIs, and ICT) trained on private augmented datasets.

\subsection{4.3\quad Ablation Study}
In this section, we first study the effectiveness of our two information losses, \emph{i.e.,} Local Information Loss $\mathcal{L}_{\mathrm{LIL}}$ and Global Information Loss $\mathcal{L}_{\mathrm{GIL}}$. We then explore the impact of local feature quantity within the local information block.

\begin{figure*}[!ht]
	\centering
        \includegraphics[width=0.85\linewidth]{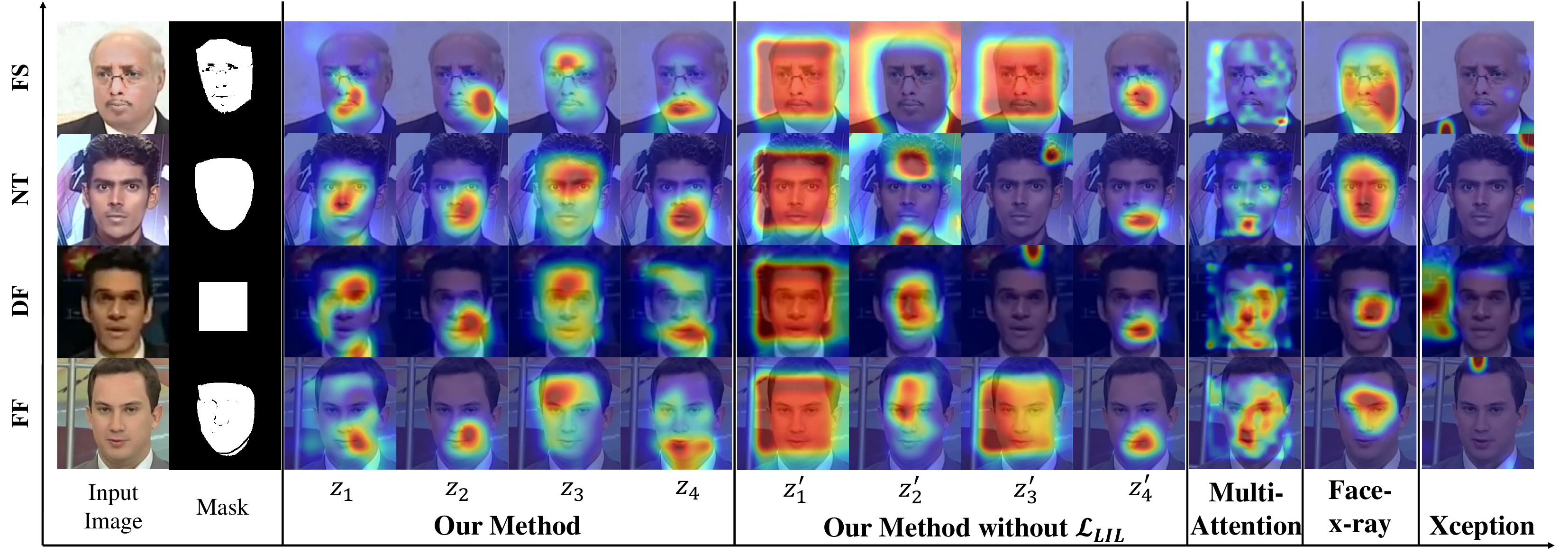}
 \caption{Visual examples of our method on various types of forgery methods within FF++ (C23), \textit{i.e.,} Deepfakes (DF), Face2Face (FF), FaceSwap (FS) and NeuralTextures (NT). Comparison between our method with and without $\mathcal{L}_{LIL}$, Multi-Attentional, Face-x-ray, and Xception. }
	\label{fig:cam} 
\end{figure*}
\begin{table}[!ht]
\centering
\small
\begin{tabular}{ccccccc}
\toprule
\multirow{2}{*}{ID} & \multicolumn{2}{c}{Loss} & \multicolumn{2}{c}{FF++ (C23)} & \multicolumn{2}{c}{CD2} \\ \cmidrule(r){2-3} \cmidrule(r){4-5}  \cmidrule(r){6-7}
  & $\mathcal{L}_{\mathrm{LIL}}$ & $\mathcal{L}_{\mathrm{GIL}}$ & ACC$\uparrow$ & AUC$\uparrow$   & ACC$\uparrow$ & AUC$\uparrow$   \\ \midrule
1 &  -   &  \Checkmark   & 94.26    & 0.979 & 78.79    & 0.840 \\ 
2 &  \Checkmark   & -    & 94.28         &   0.977    &     76.96     &  0.827     \\ 
3 &  -   &  -   & 93.53    & 0.966 & 77.29    & 0.816 \\ 
4 &  \Checkmark   &  \Checkmark   & \textbf{94.98}    & \textbf{0.983} & \textbf{80.70}    & \textbf{0.864} \\ \bottomrule
\end{tabular}
\caption{Ablation study of the proposed $\mathcal{L}_{\mathrm{LIL}}$ and $\mathcal{L}_{\mathrm{GIL}}$ for our method. We show frame-level ACC (\%) and AUC training on FF++ (C23) and testing on Celeb-DF-V2 (CD2). The bold mark best performance.}
\label{tab:as_loss}
\end{table}

\begin{figure}[!ht]
	\centering
\subfloat{
	\includegraphics[width=0.95\linewidth]{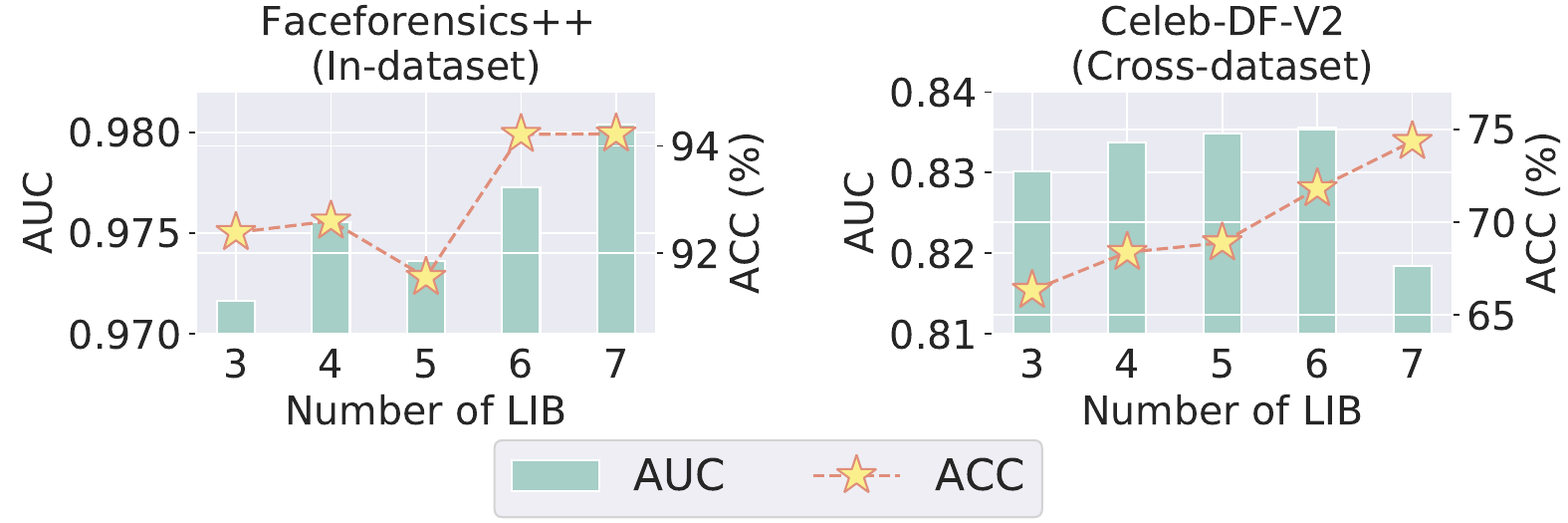}}

	\caption{In-dataset and cross-dataset performance effects within different numbers of LIBs.   We train models on FF++ (C23) with 10 epochs and test them on Celeb-DF-V2. }
	\label{fig:LIB_number} 
\end{figure}



\textbf{Ablation Study on Information Losses}\quad We study the effects of removing $\mathcal{L}_{LIL}$ and $\mathcal{L}_{GIL}$ in our method. We train models on FF++ (C23) and test them on CD2. Tab.~\ref{tab:as_loss} demonstrates the results of the ablation study on the proposed two information losses. Clearly, we see that $\mathcal{L}_{LIL}$ and $\mathcal{L}_{GIL}$ play key roles in performance improvement over in-dataset and cross-dataset settings. The AUC improvement of using the proposed losses is more critical in \emph{cross-dataset} than \emph{in-dataset} settings. This empirical evidence suggests that incorporating the proposed losses may lead to extracting broader clues. Quantitatively, $\mathcal{L}_{LIL}$ and $\mathcal{L}_{GIL}$ have a dominant contribution to our method, with AUC on FF++ improving from 0.966 (without both) to 0.983 (with both) and AUC on CD2 from 0.816 to 0.864. The absence of either loss will bring about a significant drop in model performance. 

\textbf{Ablation Study on local information block}\quad We first investigate the effect of varying the number of local information blocks (LIB), \emph{i.e.} local feature quantity. We train models on FF++ (C23) and test them on CD2, and report the frame-level AUC and ACC. The number of LIB is varied from three to seven, while other hyper-parameters are fixed. Fig.~\ref{fig:LIB_number} shows the results for different numbers of LIB. We observe that both in-dataset and cross-dataset performance improve with an increasing number of LIB increases. However, when the number of LIBs becomes excessively high ($n=7$), the model's generalization performance experiences a significant decline. This aligns with our intuition, as gradually augmenting the number of LIBs enlarges the number of trainable network parameters, directly affecting the in-dataset performance. Simultaneously, this expansion results in a rise in local feature quantity, contributing to the enhancement of the model's generalization performance. Nevertheless, as the number of LIBs continues to rise, an overabundance of parameters induces model overfitting, ultimately diminishing the model's capacity for generalization. 


\subsection{4.4\quad Visualization}

To further assess the model interpretability and the efficacy of the Local Information Loss $\mathcal{L}_{LIL}$, we visualize four samples subjected to various forgery methods on FF++. We apply Grad-CAM~\cite{selvaraju2017grad} for representation visualization. As shown in Fig.~\ref{fig:cam}, our approach offers several noteworthy insights. Firstly, it becomes evident that our method excels in extracting more forgery clues. Other detection techniques fixate on specific regions, disregarding subtle cues present elsewhere. This leads to confined regions of focus for detection. The second insight reveals our method focuses on different forgery regions with little overlap. It provides evidence that the orthogonality within extracted local representations. Specifically, our method identifies manipulated cues in the nose, cheek, forehead, and mouth, corresponding to $z_1$ through $z_4$ respectively. In contrast, results without $\mathcal{L}_{LIL}$ depict local representations possess an imbalanced capacity to signify forgery features. While some local representations contain ample information ($z_1'$), others offer duplicated ($z_3'$) or scanty ($z_2'$ and  $z_4'$) forgery-related clues. 


\subsection{4.5\quad Limitations}
{Our method is a purely data driven approach relying on information theoretic constraints to search for forgery clues. For some classes or forgeries, employing prior knowledge as guidance could be more optimal. For future work, we seek to incorporate heuristic guidance into our model, which could further boost performance and interpretability.}

\section{5\quad Conclusion}
{In this paper, we propose an information bottleneck based framework for deepfake detection, which aims to extract broader forgery clues. In this context, we derive local information losses to obtain task-related independent local features. We further theoretically analyze the global information objective to aggregate local features into a sufficient and purified global representation for classification.  Extensive experiments demonstrate that our method achieves state-of-the-art \emph{in-dataset} and \emph{cross-dataset} performance on five benchmark datasets, indicating its potential as a reliable solution for deepfake detection in various real-world scenarios.}

\section*{Acknowledgements}
This work is partially supported by Zhejiang Provincial Natural Science Foundation of China under Grant No. LD24F020010, the National Natural Science Foundation of China (62172359, 62372402, 61972348, 62102354), the Key R\&D Program of Zhejiang Province (2023C01217), the Fundamental Research Funds for the Central Universities (No. 2021FZZX001-27), and the Hangzhou Leading Innovation and Entrepreneurship Team (TD2020003).


\appendix
\section{Appendix}
In this section, we will first provide the proof for the theorem (the lower bound of local information objective) and the formulation derivations of the Local Information Loss $\mathcal{L}_{LIL}$ in the main text. After that, we will also present additional experimental results.

\subsection{A.\label{sec:proof} \quad Theorem Proof}

\subsubsection{Formulation.}
\begin{figure}[!ht]
    \centering
    \includegraphics[width=1\linewidth]{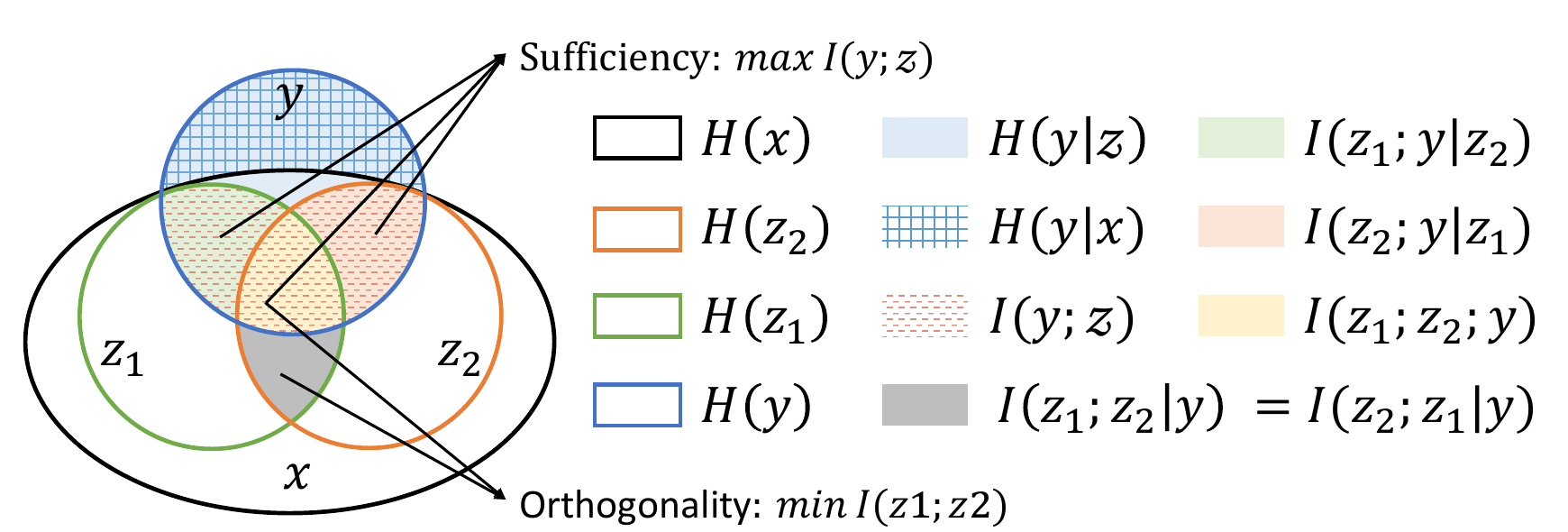}
    \caption{Information content of local representations when $n=2$.}
    \label{fig:appendix_lil}
\end{figure}

Given an input $x$ and a label $y$, $\mathcal{Z}$ serves as the joint local representation, expressed as $\mathcal{Z}=\bigoplus_i^n z_i$. The notation $\mathcal{Z} \setminus z_i \equiv z_1 \oplus \cdots \oplus z_{i-1} \oplus z_{i+1} \oplus \cdots \oplus z_n$ represents the joint probability $\mathcal{Z}$ except $z_i$. To facilitate understanding, Figure~\ref{fig:appendix_lil} illustrates the information content of local representations when $n=2$. 

Our local information objective aims to ensure two essential properties in local representations, \emph{i.e.,} \textbf{comprehensiveness and orthogonality}. Comprehensiveness entails that each local representation $z_i$ contains a sufficient amount of label-related information within $x$. To achieve this goal, we maximize the mutual information between $y$ and $\mathcal{Z}$, formulated as $\max I(y;\mathcal{Z})$. Concurrently, orthogonality requires that any pair of local representations $z_i$ and $z_j$ is orthogonal, implying that the label-related mutual information between any two representations is zero, expressed as $\min \sum_{i\neq j}^{n} I(z_i;z_j)$. 

We derive our objective further by drawing inspiration from the information bottleneck theory. The concept of information bottlenecks is proposed in \cite{tishby2000information} which attributes the robustness of a machine learning model to its ability to distill superfluous noise while retaining only useful information. Thus, the information outside the label $y$ is all superfluous for the task, such as the gray region in Figure~\ref{fig:appendix_lil}. We primarily retain the task-related information, \emph{i.e.,} $\max \sum_{i\neq j}^{n} I(z_i;z_j;y)$, and eliminate the superfluous information through the Global Information Loss $\mathcal{L}_{GIL}$. Furthermore, the two properties of local representations are in an adversarial relationship with the mutual information between $y$, $z_i$, and $z_j$, depicted as $I(y;z_i;z_j)$ (the yellow region in Figure~\ref{fig:appendix_lil}). Therefore, the final local information objective is:
\begin{equation}
    \begin{aligned}
    \max\sum_{i=1}^{n} I( z_i;y &\mid \mathcal{Z} \setminus z_i ).
\end{aligned}
\end{equation}

\subsubsection{Theorem.}
To guarantee both comprehensive and orthogonality of local representations, the local information objective has a lower bound:
\begin{equation}
\begin{aligned}
 \sum_{i=1}^{n} I( z_i;y\mid \mathcal{Z} \setminus z_i ) \geq \sum_{i=1}^{n} D_{KL}\left[\mathbb{P}_z \| \mathbb{P}_{z \setminus z_i}\right],
\end{aligned}
\label{eq:lower_bound}
\end{equation}

\subsubsection{Proof.}
In accordance with the definition of conditional mutual information~\cite{wyner1978definition}, given three random variables $X$, $Y$, and $Z$, the conditional mutual information can be expressed as follows:
\begin{equation}
    \begin{aligned}
    I(X ; Y \mid Z)=\sum_{ X,Y,Z} p(X, Y, Z) \log \frac{p(Z) p(X, Y, Z)}{p(X, Z) p(Y, Z)}.
    \end{aligned}
\end{equation}
Therefore, our local objective can be expressed as:
\begin{equation}
\begin{aligned}
& \sum_{i=1}^{n} I( z_i;y \mid \mathcal{Z} \setminus z_i ) \\
= &\sum_{i=1}^{n} \sum_{y,\mathcal{Z}} p(y,\mathcal{Z}) \log \frac{p(\mathcal{Z} \setminus z_i) p(y,\mathcal{Z})}{p(\mathcal{Z}) p(y, \mathcal{Z} \setminus z_i)}\\
= &\sum_{i=1}^{n} \left\{ \underbrace{\sum_{y,\mathcal{Z}}p(y,\mathcal{Z})\log \frac{p(\mathcal{Z} \setminus z_i)} {p(\mathcal{Z})}}_{Q_1} \right.\\
&\left.\quad \quad \quad \quad \quad \quad + \underbrace{\sum_{y,\mathcal{Z}}p(y,\mathcal{Z})\log \frac{p(y,\mathcal{Z})}{p(y, \mathcal{Z} \setminus z_i)}}_{Q_2} \right\}.
\end{aligned}
\end{equation}

Based on Bayes' theorem, \emph{i.e.,} $p(X,Y)=p(X)p(Y|X)$, the term $Q_1$ can be expanded as follows:
\begin{equation}
\begin{aligned}
Q_1 = & \sum_{y,\mathcal{Z}} p(y,\mathcal{Z})\log \frac{p(\mathcal{Z}\setminus z_i)}{p(\mathcal{Z}\setminus z_i,z_i)}\\
= & \sum_{y,\mathcal{Z}} p(y,\mathcal{Z})\log \frac{1}{p(z_i|\mathcal{Z}\setminus z_i)}\\
= & \sum_{y,\mathcal{Z}} p(y,\mathcal{Z})\log 1-\sum_{y,\mathcal{Z}} p(y,\mathcal{Z})\log p(z_i|\mathcal{Z}\setminus z_i)\\
= &-\sum_{y,\mathcal{Z}} p(y|\mathcal{Z})p(\mathcal{Z})\log p(z_i|\mathcal{Z}\setminus z_i)\\
= & -\sum_{y} p(y|\mathcal{Z}) \sum_{\mathcal{Z}}p(\mathcal{Z})\log p(z_i|\mathcal{Z}\setminus z_i),
\end{aligned}
\label{eq:q_1}
\end{equation}
According to the definition of conditional entropy, \emph{i.e.,} $H(X|Y) = -\sum_{X,Y} p(X,Y)\log p(X|Y),$ Eq.~\ref{eq:q_1} becomes,
\begin{equation}
\begin{aligned}
Q_1 = \sum_{y} p(y|\mathcal{Z}) H(z_i|\mathcal{Z}\setminus z_i).
\end{aligned}
\end{equation}
Referring to the definition of KL-divergence, \emph{i.e.,} $D_{KL} = \sum_{X,Y} p(X)\log \frac{p(X)}{p(Y)},$ the term $Q_2$ thus becomes:
\begin{equation}
\begin{aligned}
Q_2=D_{K L}\left[p(y \mid \mathcal{Z}) \| p\left(y \mid \mathcal{Z} \setminus z_i\right)\right],
\end{aligned}
\end{equation}

Overall, we have:
\begin{equation}
\begin{aligned}
\sum_{i=1}^{n} I( z_i &;y \mid \mathcal{Z} \setminus z_i ) = \sum_{i=1}^{n}  \sum_{y} p(y|\mathcal{Z}) H(z_i|\mathcal{Z}\setminus z_i) \\
&  + \sum_{i=1}^{n} D_{K L}\left[p(y \mid \mathcal{Z}) \| p\left(y \mid \mathcal{Z} \setminus z_i\right)\right] .
\end{aligned}
\end{equation}
Due to the non-negativity of information entropy and probability, it follows that $Q_1 \geq 0$ . As a result, we deduce the lower bound:
\begin{equation}
\begin{aligned}
 \sum_{i=1}^{n} I( z_i;y\mid \mathcal{Z} \setminus z_i ) \geq \sum_{i=1}^{n} D_{KL}\left[\mathbb{P}_z \| \mathbb{P}_{z \setminus z_i}\right],
\end{aligned}
\label{eq:final}
\end{equation}
where $\mathbb{P}_{\mathcal{Z} \setminus z_i} = p \left( y\mid \mathcal{Z} \setminus z_i \right), \mathbb{P}_\mathcal{Z} = p \left(y\mid \mathcal{Z} \right)$. Given that the KL divergence ranges from 0 to infinity, we employ the exponential function to convert the objective from maximization to minimization. Consequently, we can formalize the local information objective as follows:
\begin{equation}
\begin{aligned}
\min _\theta \exp \left( -\sum_{i=1}^{n} D_{KL}\left[\mathbb{P}_z \| \mathbb{P}_{z \setminus z_i}\right]\right),
\end{aligned}
\end{equation}
where $\theta$ represents the model parameters of our local disentanglement module. When $\exp \left( -\sum_{i=1}^{n} D_{KL}\left[\mathbb{P}_z \| \mathbb{P}_{z \setminus z_i}\right]\right)$ tends toward zero, $\sum_{i=1}^{n} D_{KL}\left[\mathbb{P}_z \| \mathbb{P}_{z \setminus z_i}\right]$ tends towards infinity, and then $\sum_{i=1}^{n} I( z_i;y\mid \mathcal{Z} \setminus z_i )$ also tends to infinity. Thus, we arrive at the optimal comprehensiveness and orthogonality of local representations, which proves the theorem in Eq.~\ref{eq:lower_bound}.

\subsection{B.\quad Evaluation} \label{sec:eval}

\begin{figure*}[!ht]
	\centering
        \includegraphics[width=\linewidth]{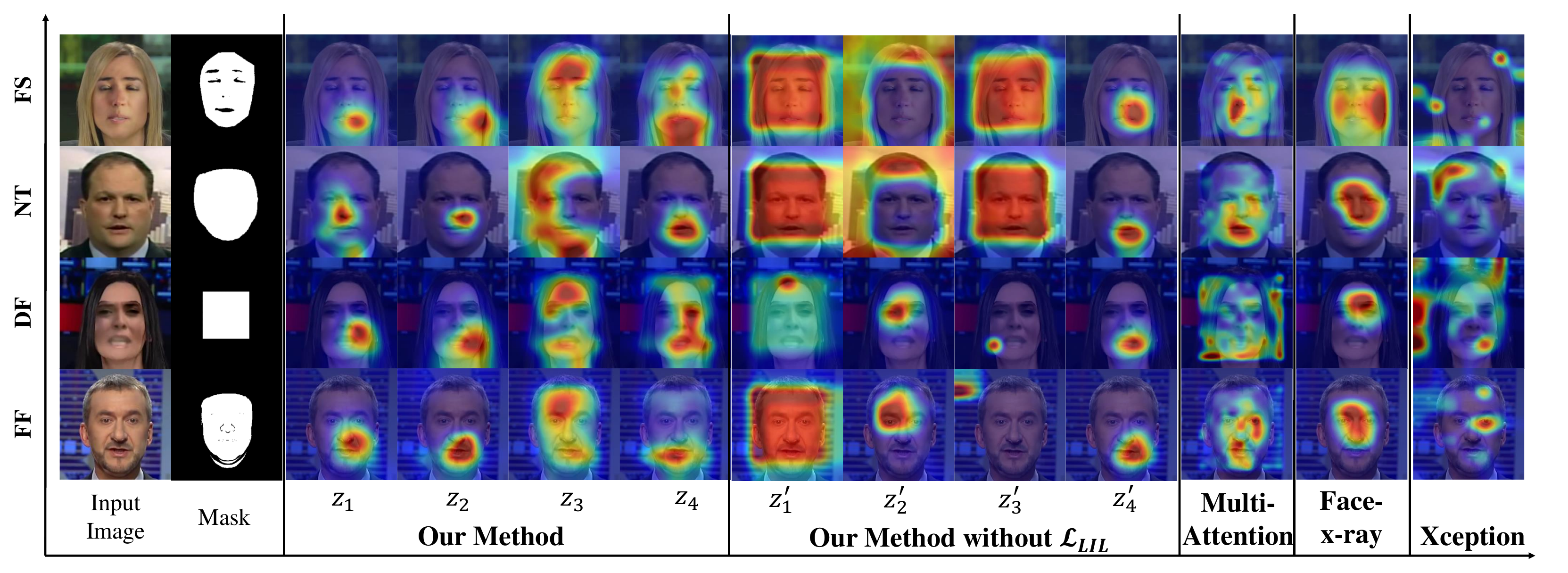}
 \caption{More visual examples of our method on various types of forgery methods, \textit{i.e.,} Deepfakes (DF), Face2Face (FF), FaceSwap (FS) and NeuralTextures (NT). Comparison between our method with and without $\mathcal{L}_{LIL}$, Multi-Attentional~\cite{zhao2021multi}, Face-x-ray~\cite{li2020face}, and Xception~\cite{rossler2019faceforensics++}. Our method extracts more forgery clues and focuses on different regions with little overlap. Local Information Loss $\mathcal{L}_{LIL}$ is proven to guarantee the orthogonality between multiple local features and the comprehensiveness of relevant label information. In contrast, Multi-Attentional, Face-x-ray, and Xception all grasp fewer regions. }
	\label{fig:cam_app} 
\end{figure*}

\subsubsection{Implementation details.} 
In data pre-processing, we crop all videos at an interval of five frames and then use the SOTA face extractor RetinaFace~\cite{deng2020retinaface} to obtain face regions for training. In addition, we oversample pristine videos for balancing training datasets. We also apply mixup~\cite{zhang2017mixup} and other common augmentation techniques such as rotation, flip, brightness, and contrast. The learning rate $lr$ is initialized to $5\times 10^{-4}$ and decays by 50\% every five epochs. We use the Adam optimizer~\cite{kingma2014adam} and adopt cosine annealing~\cite{loshchilov2016sgdr} as the scheduler.

\begin{table}[!ht]
\centering
\begin{tabular}{lcccc}
\toprule
                & \multicolumn{2}{c}{FF++ (C23)} & \multicolumn{2}{c}{CD2} \\ \cmidrule(r){2-3} \cmidrule(r){4-5}
                & ACC$\uparrow$            & AUC$\uparrow$            & ACC$\uparrow$                   & AUC$\uparrow$  \\ \midrule
MobileNet-V1 & 90.85 & {0.962} & 70.36                     & 0.858                     \\
EfficientNet-B0 & 93.71          & 0.982          & {72.32}  &  0.863    \\
ResNet-34    & {94.98} & {0.983} & {80.70} & {0.864} \\ \bottomrule
\end{tabular}
\caption{Effect of different backbones of the local information block. We engage in different backbones including MobileNet, EfficientNet, and ResNet. We train our method on FF++ (C23) and test it on Celeb-DF-V2 (CD2). Here we report the frame-level ACC (\%) and AUC. }
\label{tab:LIB_backbone}

\end{table}

\subsubsection{The backbone type for the local information block.}
We study the influence of different backbone alternatives in the local information block (LIB). We select three popular backbone networks for our LIB, \emph{i.e.}, MobileNet~\cite{howard2017mobilenets}, EfficientNet~\cite{tan2019efficientnet}, and ResNet~\cite{he2016deep}. We train the model on FF++ (C23) and test it on Celeb-DF-V2. Results are shown in Tab.~\ref{tab:LIB_backbone}. The performance of our method is stable with respect to different backbones, retaining a similar level of in-dataset and cross-dataset performance. The model with ResNet-34 as LIB obtains the best performance in both in-dataset and cross-dataset settings. Compared with MobileNet-V1 and EfficientNet-B0, ResNet-34 has the largest number of parameters, which may be the reason for its better performance.

\subsubsection{Visualization.}
We further provide more visualization results to demonstrate the interpretability of our method. We apply Grad-CAM for visualizing local representations of our method and compare our method with our method without Local Information Loss $\mathcal{L}_{LIL}$, and other detection approaches. Specifically, we select state-of-the-art methods, \emph{i.e.,} Multi-Attentional~\cite{zhao2021multi}, Face x-ray~\cite{li2020face} and Xception~\cite{rossler2019faceforensics++}. We randomly select four samples from four forgery methods within FF++ (C23) dataset for visualization, respectively. Results are demonstrated in Figure~\ref{fig:cam_app}. We observe that our method obtains broader clues and disentangles forgery regions, where each local representation is concentrated in an effective region and has little intersection with other local representations (regions). In contrast, for results without $\mathcal{L}_{LIL}$, local representations focus on duplicated regions, and their abilities to signify forgery features are unbalanced. Some local representations own sufficient information while some scarcely provide forgery clues. Furthermore, other detection methods focus on fewer regions. Specifically, Xception and Face-x-ray fall into the deficiency of a small region and ignore the other regions. They are unable to give fair attention to every forgery region. Multi-Attentional is aware of the importance of local subtle forgery clues, but it cannot disentangle local regions under the theory guarantee and the entire attention scope has not become essentially broader. These methods fail to consider more comprehensive clues, which translates to lower generalization capacity.

\subsubsection{Scalability.}
We collect 24 in-the-wild deepfake videos from YouTube to further evaluate the scalability. The models are trained on FF++ and tested on in-the-wild deepfakes. Our method achieves 0.728 AUC, outperforming state-of-the-art methods such as Xception (0.579).

\subsubsection{Robustness.} 
We add an evaluation on the robustness of our detection method. Due to space limitations, we present the results for three types of perturbations in Tab.~\ref{tab:robutness}. The perturbations are \emph{contrast, saturation}, and \emph{compression}. Our method shows robust performance against different image degradations.

\begin{table}[!ht]
\centering
\scriptsize
\begin{threeparttable}
\begin{tabular}{cccccc} 
\hline
Perturbations & Level 1 & Level 2 & Level 3  & Level 4 & Level 5\\ \hline
Normal &  0.983   & 0.983   & 0.983& 0.983& 0.983\\
Contrast &  0.977 &	0.977&	0.976&	0.975&	0.967  \\
Saturation & 0.979 &	0.979 &	0.978&	0.978&	0.978 \\
Compression&   0.972& 	0.966	& 0.952	& 0.939	& 0.918\\
Average & 0.976	&0.974	&0.969	&0.964&	0.955 \\ 
\hline
\end{tabular}
\begin{tablenotes} 
\item[--] Frame-level AUC w.r.t. five severity levels for each degradation form.
\end{tablenotes}
\end{threeparttable}
\vspace{-0.25cm}
\caption{Robustness to different image degradations.}
\label{tab:robutness}
\end{table}

\subsubsection{Runtime analysis.}
We add a runtime evaluation, listed in Tab.~\ref{tab:runtime}. We can observe that: (1) The training time increases as $n$ (number of local features extracted) increases. (2) Our method achieves better accuracy while consuming similar training time. 

\begin{table}[!ht]
\centering
\scriptsize
\setlength\tabcolsep{2.5pt}
\begin{threeparttable}
\begin{tabular}{lcccccc} \hline
Method     & \begin{tabular}[c]{@{}c@{}}FLOPs\\ (G)\end{tabular} 
& \begin{tabular}[c]{@{}c@{}}MACs\\ (G)\end{tabular}
& \begin{tabular}[c]{@{}c@{}}Params\\ (M)\end{tabular}
& \begin{tabular}[c]{@{}c@{}}Time \\ (GPU:h)\end{tabular}
& \begin{tabular}[c]{@{}c@{}}AUC \\ (FF++)\end{tabular} 
& \begin{tabular}[c]{@{}c@{}}AUC \\ (CD2)\end{tabular}
\\ \hline
Xception   & 16.78     & 8.36     & 20.81      & 2:7.67   & 0.963	&0.778    \\
Face X-ray & 21.30     & 10.61    & 30.95      & 4:11.80   &0.874	&0.742   \\ \hline
Ours ResNet34 ($n=3$) & 22.02     & 10.99    & 66.16      & 2:7.67  & 0.972&0.830      \\
Ours ResNet34 ($n=4$) & 29.36     & 14.66    & 87.97      & 2:10.17 & 0.976	& 0.834      \\
Ours ResNet34 ($n=5$) & 36.70     & 18.32    & 109.78     & 2:11.67  & 0.974	& 0.835    \\
Ours ResNet34 ($n=6$) & 44.04     & 21.98    & 131.59     & 3:13.33    &0.977	& 0.836  \\
Ours ResNet34 ($n=7$) & 51.38     & 25.65    & 153.40     & 3:15.40  &0.980	&0.818       \\ \hline
Ours MobileNet-V1 ($n=4$)&	4.61&	2.28&	22.08&	2:8.25 & 0.962	&0.858 \\
Ours EfficientNet-B0 ($n=4$)&	3.15&	1.54&	27.47&	4:10.67 &0.982	&0.863 \\ \hline
\end{tabular}
\begin{tablenotes} 
\item[--] The ‘Time’ column shows the number of GPUs used and training time (in hours). 
\item[--] For a fair comparison, all models are trained for 20 epochs with a batch size of 256 on the FF++ dataset. 
\end{tablenotes}
\end{threeparttable}
\caption{Runtime analysis. }
\label{tab:runtime}
\end{table}

\subsubsection{Cropping evenly method.}
There is a possible naive method that pre-defines several non-overlapping regions (cropping evenly) and then separately extracts and aggregates their features.
We have tried this cropping evenly method and compared it with the proposed method. Its in-dataset AUC is 0.959 (vs ours 0.983) and its cross-dataset AUC is 0.798 (vs ours 0.864).  Our method consistently outperforms this method. We conjecture the reasons are: (i) Segmenting the image into multiple regions may destroy the integrity of the forgery clues.  (ii) The distribution of various regions of forgery clues may be uneven. 

\end{document}